\documentclass[conference]{IEEEtran}
\IEEEoverridecommandlockouts
\usepackage{cite}
\usepackage{amsmath,amssymb,amsfonts}
\usepackage{algorithmic}
\usepackage{graphicx}
\usepackage{subcaption}
\usepackage{textcomp}
\usepackage{xcolor}
\usepackage{balance}
\usepackage{lipsum}
\usepackage{fancyhdr}
\def\BibTeX{{\rm B\kern-.05em{\sc i\kern-.025em b}\kern-.08em
    T\kern-.1667em\lower.7ex\hbox{E}\kern-.125emX}}
\begin{document}

\title{Exciting Contact Modes in Differentiable Simulations for Robot Learning}

\author{\IEEEauthorblockN{Hrishikesh Sathyanarayan}
\IEEEauthorblockA{\textit{Dept. of Mechanical Engineering} \\
\textit{Yale University}\\
New Haven, USA \\
hrishi.sathyanarayan@yale.edu}
\and
\IEEEauthorblockN{Ian Abraham}
\IEEEauthorblockA{\textit{Dept. of Mechanical Engineering} \\
\textit{Yale University}\\
New Haven, USA \\
ian.abraham@yale.edu}
\thanks{This material is based upon work supported by the National Science Foundation under award NSF FRR 2238066. Any opinions, findings, and conclusions or recommendations expressed in this material are those of the authors and do not necessarily reflect the views of the National Science Foundation.}
}

\maketitle
\thispagestyle{fancy}

\begin{abstract}
In this paper, we explore an approach to actively plan and excite contact modes in differentiable simulators as a means to tighten the sim-to-real gap. 
We propose an optimal experimental design approach derived from information-theoretic methods to identify and search for information-rich contact modes through the use of contact-implicit optimization. 
We demonstrate our approach on a robot parameter estimation problem with unknown inertial and kinematic parameters which actively seeks contacts with a nearby surface.
We show that our approach improves the identification of unknown parameter estimates over experimental runs by an estimate error reduction of at least $\sim 84\%$ when compared to a random sampling baseline, with significantly higher information gains.
\end{abstract}


\vspace{-0.5em}
\section{Introduction}

    Differentiable simulators are quintessential for many model-based control and model-based reinforcement learning methods~\cite{6386109,howell2022predictivesamplingrealtimebehaviour,NEURIPS2018_842424a1,hu2020difftaichidifferentiableprogrammingphysical}, but the effectiveness of such simulations are limited as a consequence of the sim-to-real gap and parameter estimation inaccuracy. 
    Prior work has demonstrated accurate identification of unknown parameters by leveraging contact interactions with the environment~\cite{nima, MI_review}. 
    However, exciting and planning contact interactions is challenging due to the sparse and non-smooth nature of contact. 
    In this work, we propose a method that plans \emph{meaningful} contact interactions for robots to obtain information-rich data that facilitates improved parameter learning.
    
    In order to search for rich contact data for effective parameter learning, we require the identification of key contact interactions and a way to plan control inputs to explore for contacts. 
    To achieve this, we propose an experimental design approach \cite{atanasov2013information,fedorov2010optimal,abraham_thesis} that is based on contact-implicit optimization that optimizes contact modes which maximizes a contact-aware Fisher information  \cite{wilson_sac, wilson_fishermax} which directly reduces a lower-bound on parameter uncertainty \cite{Rao_1947,abraham_thesis}.
    Our approach integrates contact planning~\cite{directposa,1631739,Yuntarticle,10.1007/978-1-4020-6332-9_19,liu2024softmacdifferentiablesoftbody, Manchester_Doshi_Wood_Kuindersma, nima} with the contact-aware information metric to facilitate smoothness in the computation of gradients that are useful for down-stream learning tasks. 
    We show that our approach is able to search for contact modes that maximizes information which improves learning of unknown parameters of interest, thus guiding and enhancing the robot learning process.

\begin{figure*}[t]
    \centering
    \begin{subfigure}[b]{0.24\textwidth} 
        \centering
        \includegraphics[width=\textwidth]{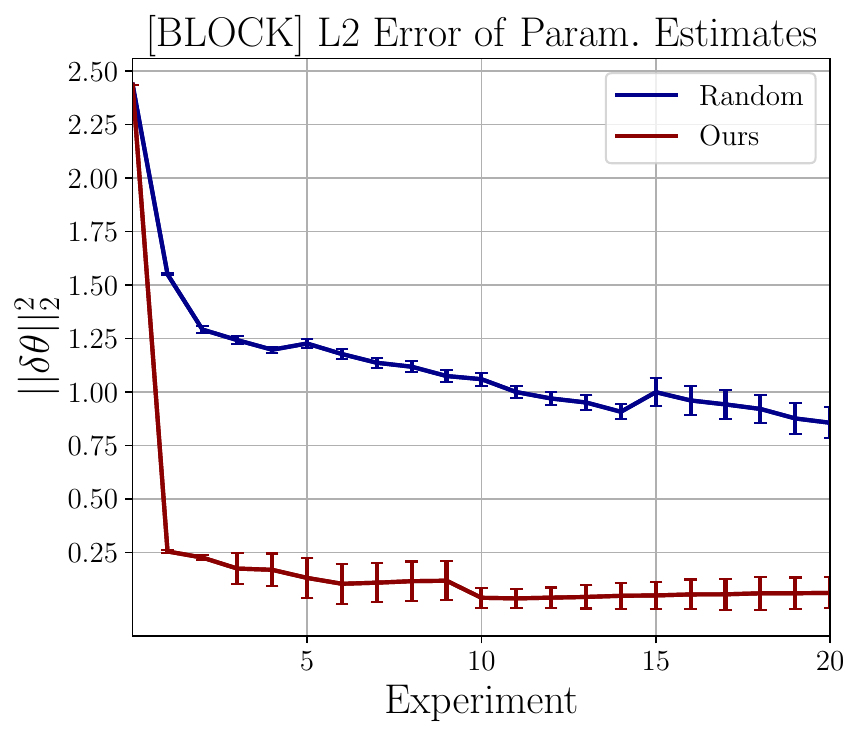}
        \caption{}
        \label{fig:block_param}
    \end{subfigure}
    \begin{subfigure}[b]{0.24\textwidth} 
        \centering
        \includegraphics[width=\textwidth]{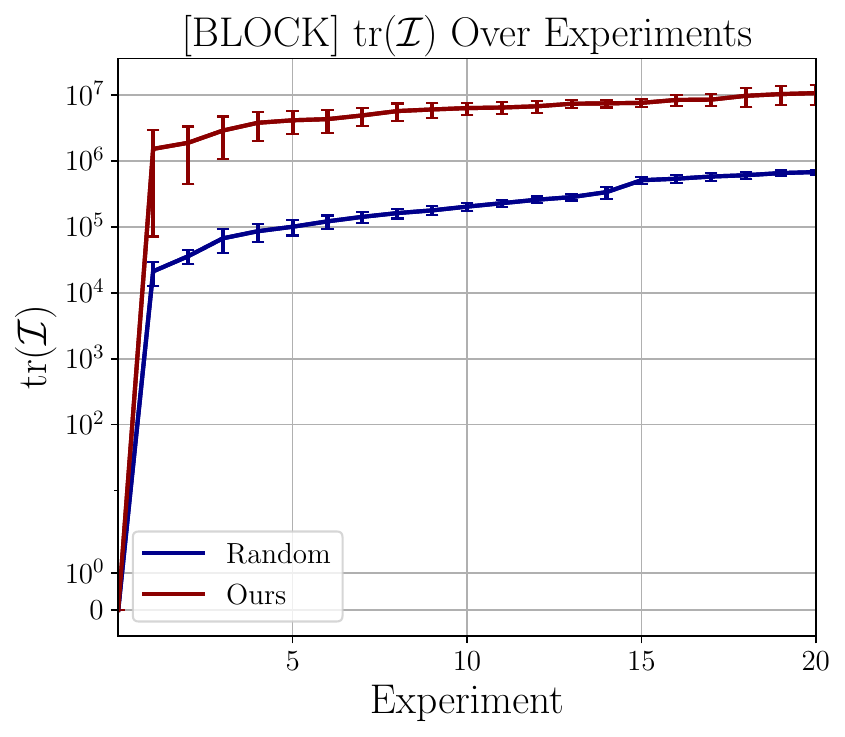}
        \caption{}
        \label{fig:block_fim}
    \end{subfigure}
    \begin{subfigure}[b]{0.24\textwidth} 
        \centering
        \includegraphics[width=\textwidth]{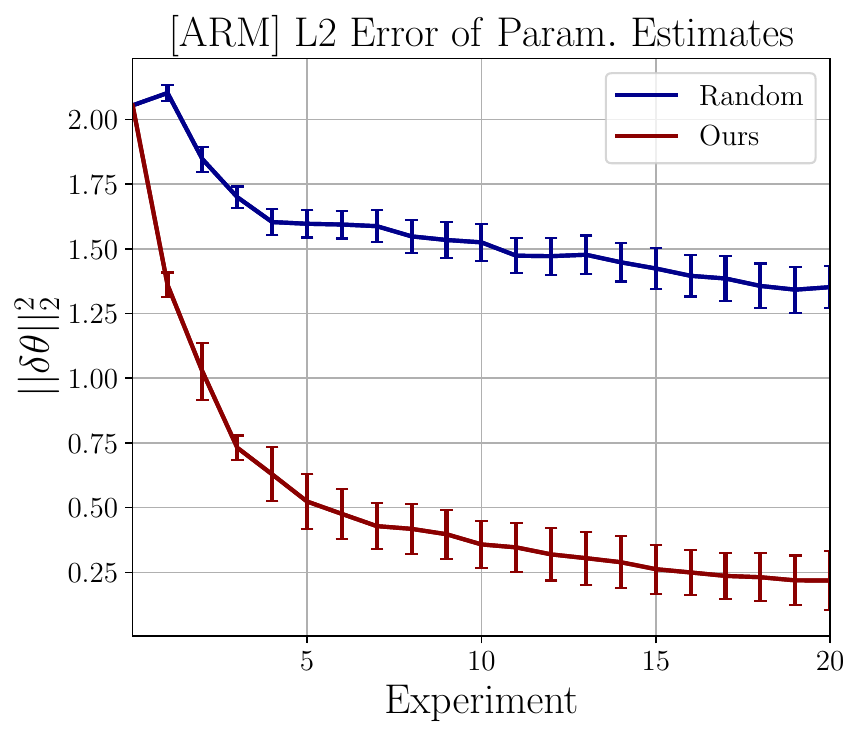}
        \caption{}
        \label{fig:oc_param}
    \end{subfigure}
    \begin{subfigure}[b]{0.24\textwidth} 
        \centering
        \includegraphics[width=\textwidth]{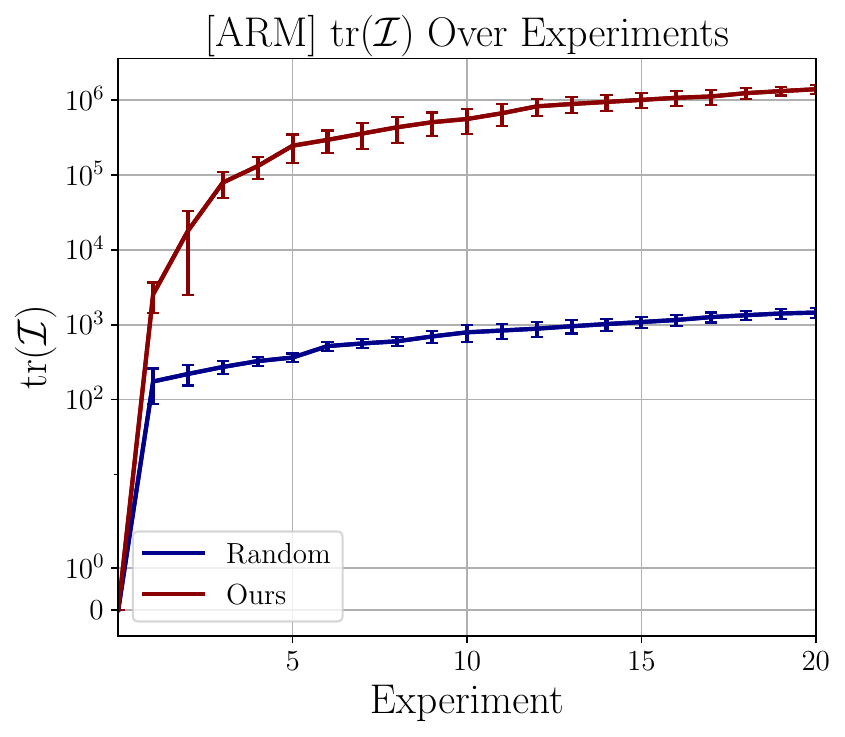}
        \caption{}
        \label{fig:oc_fim}
    \end{subfigure}
    \caption{Results for the planar block throwing and three-link planar arm evaluated over 20 experiments. Compared to uniform random sampling methods, we observe $\sim 97 \%$ and $\sim 84 \%$ parameter estimates error reduction for the (a) block and (c) arm experiments, respectively, and we observe higher information gains over experiments for both the (b) block and (d) arm.}
    \label{fig:results_over_experiments}
\end{figure*}

\vspace{-0.5em}
\section{Methods}
\label{sec:methods}

    The underlying problem that we are interested in is the contact-aware maximum likelihood problem,
    \begin{align*}\label{eq:max_like}
        &\ \ \ \ \ \ \ \ \ \ \ 
        \max_{\theta, \{ x_i, \lambda_i \}_{i=0}^N} \log \prod_{i=0}^N p(\bar{y}_i | x_i, u_i, \lambda_i, \theta)  \\ 
        \text{s.t.} & 
        \begin{cases}
            x_0, u_i \forall i \in [0,N] & \text{ (givens) } \nonumber \\ 
            y_i = g(x_i, u_i, \lambda_i, \theta) & \text{ (sensor model) } \nonumber \\ 
            x_{i+1} = f(x_i, u_i, \lambda_i, \theta) & \text{ (dynamics model) }\nonumber \\ 
            \lambda_i \in \mathcal{C}(x_i, u_i, \theta) & \text{ (contact constraints) } \nonumber \\ 
            \theta \in \Theta \nonumber & \text{ (feasible parameter) }
        \end{cases} \\
    \end{align*}
    where $\{ \bar{y}_i \}_{i=0}^N$ are sensor readings from running open loop controls $u_i \in \mathcal{U}$ with initial condition $x_0$.
    This problem optimizes over physics parameters $\theta$ and states $x_i \in \mathcal{X}$ and contacts $\lambda_i \in \mathcal{C}(x_i, u_i, \theta)$.
    Here, $f$ are the simulation dynamics that takes as input the state, control, feasible contact forces, and current parameter estimates, and returns the subsequent state $x_{i+1}$, and $p(y | x, u, \lambda, \theta)$ is the measurement likelihood model (with mean $g(x, u, \lambda, \theta)$) that captures the uncertainty in the measurements. 
    Our goal is to produce a control sequence $u_i \forall i \in [0,N]$ that generates a dataset $\{ \bar{y}_i \}_{i=0}^N$ that excites the most informative contacts that improves learning $\theta$. 
    
    To achieve this, we leverage the empirical Fisher information~\cite{AF_Emery_1998} to quantify information-richness, which is a lower-bound on the variance of the posterior of the maximum likelihood problem as 
        \begin{equation}\label{eq:FIM_general}
               \mathcal{I}(\tau | \theta) 
                \approx \sum_{i=0}^N  \xi_i \xi_i^\top \ge \text{var}\left[ \theta^\star \right]^{-1}
        \end{equation}
    where $\xi_i = \nabla_\theta \log p(y_i\ |\  x_i,u_i, \lambda_i, \theta)$, $\theta^\star$ is the posterior, $\tau = [(x_0,u_0,\lambda_0), \ldots, (x_N,u_N,\lambda_N)]$ is the trajectory. 
    Maximizing optimality conditions on the Fisher information matrix can often yield more meaningful and diverse datasets that enables more effective learning~\cite{CHIN2020989,wagenmaker2020activelearningidentificationlinear}.
    Using the Fisher information, we pose the following optimization problem: 
    \begin{equation}
        \label{eq:exp_learning_opt}
        \begin{aligned}
                \max_{\{x_i,u_i, \lambda_i\}_{i=0}^N} \text{tr} \left[ \mathcal{I}( \tau | \theta) \right] & &
                \text{s.t. } &
                \begin{cases}
                    x_0 \text{     (given)}\\
                    x_{i+1} = f(x_i,u_i,\lambda_i,\theta),\nonumber \\
                    \lambda_i \in \mathcal{C}(x_i, u_i, \theta) \nonumber
                \end{cases}
        \end{aligned}
    \end{equation}
    and $\theta$ is the current estimate of the parameters of the simulator. 
    What makes our approach novel is the explicit consideration of contact forces $\lambda$ in the Fisher information maximization. 
    
\begin{figure}
    \centering
    \begin{subfigure}[b]{0.22\textwidth} 
        \centering
        \includegraphics[width=\textwidth]{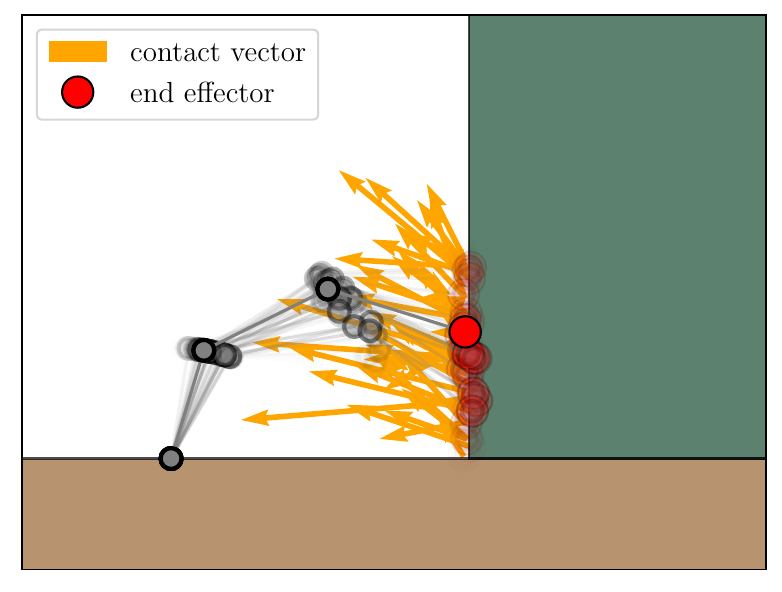}
        \caption{}
        \label{fig:oc_ex}
    \end{subfigure}
    \begin{subfigure}[b]{0.22\textwidth} 
        \centering
        \includegraphics[width=\textwidth]{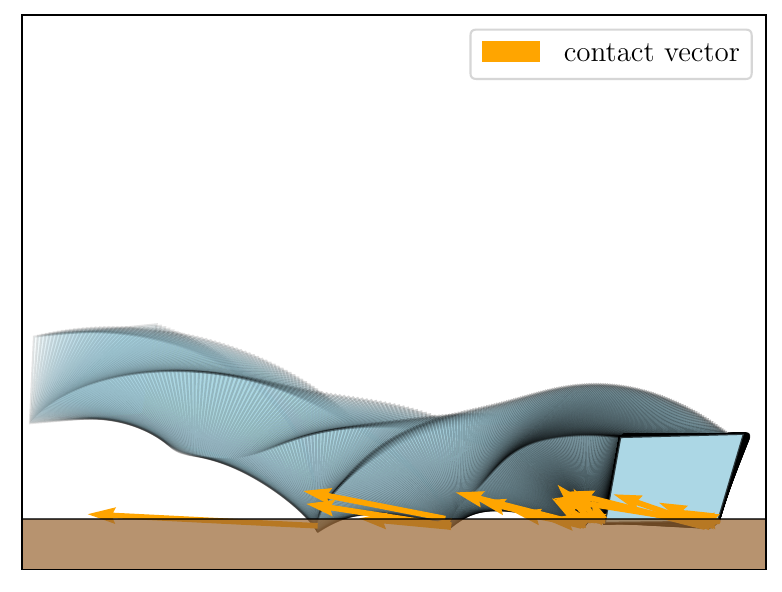}
        \caption{}
        \label{fig:block_ex}
    \end{subfigure}
    \caption{Arm (a) and block throwing (b) example contact-aware experiment via our proposed approach.}
    \label{fig:experiments}
\end{figure}

\vspace{-0.5em}
\section{Results and Discussion}
\label{sec:result}

    We show the effectiveness of our experimental design approach in two differentiable simulations by comparing with uniform random sampling approaches. 
    Our approach utilizes the choice of two distinct sensor models, described as functions of the parameter estimates $\theta$, contact forces $\lambda$, robot state $\mathbf{x}=[\mathbf{q^\top}, \mathbf{\dot{q}^\top}]^\top$ and joint torque control $\mathbf{u}$, where $\mathbf{q}$ and $\mathbf{\dot{q}}$ are joint positions and velocities, respectively. 
    We define the contact Jacobian $\mathbf{J_c(q, \theta)}$ as the projection of the contact forces at the contact frame to joint torques, and the sensor Jacobian $\mathbf{J_g}$ as a projection of velocities from the world frame to the sensor frame. 
    Furthermore, we define $\mathbf{M(q, \theta)}$ as the mass matrix, and $\mathbf{b(q,\dot{q})}$ as the bias.
    
    \noindent
    \textbf{Robot Parameter Estimation from Contact.} 
    We evaluate our approach on a three link planar robot arm. 
    We utilize the following contact force sensor model,

   \begin{equation}\label{eq:contact_sensor}
       \begin{split}
           g(\mathbf{x}, \mathbf{u}, \lambda, \theta) = \mathbf{\left(J_c(q,\theta) J_c^T(q,\theta)\right)^{-1}}&\mathbf{J_c(q,\theta)} (\mathbf{M(q,\theta)} \mathbf{\Ddot{q}} \\
           &+ \mathbf{b(q, \dot{q}) - u})
       \end{split}
   \end{equation}
   where $\mathbf{\Ddot{q}}$ is approximated numerically.
    
    The unknown parameters are given by the joint link inertia and the kinematic link lengths. We show in Fig. (\ref{fig:oc_ex}) the arm exciting contact rich behavior through our optimized Fisher information maximization control approach. 
    Consequently, we are able to improve parameter estimation error reduction by $\sim 84\%$ compared to a uniform random sampling based method, as shown in Fig. (\ref{fig:oc_param}).
    Observing Fig. (\ref{fig:oc_fim}), we see that our approach generates contacts that  are more information-rich with respect to the Fisher information parameters of the differentiable simulator. 
    
    

    \noindent
    \textbf{Block Throwing Parameter Estimation.} 
    We evaluate our approach for a planar block throwing system. We utilize the following  accelerometer sensor model,
    \begin{equation}
        \begin{split}
            g(\mathbf{x}, \mathbf{u}, \lambda, \theta) = \mathbf{\dot{J}_{g}\dot{q}} + \mathbf{J_{g}} (\mathbf{M(q,\theta)^{-1}}& (\mathbf{J_c^T(q,\theta)} \mathbf{\lambda} \\
            &- \mathbf{b(q, \dot{q}) + \mathbf{u}})).
        \end{split}  
    \end{equation}
    Here, because the use of a contact force sensor is generally intractable for block throwing systems, the contact forces $\lambda$ are inferred implicitly during parameter estimation. The unknown parameters are given by the block mass and geometry. We show in Fig. (\ref{fig:block_ex}) the choice of initial throw for the block exciting contact rich behavior through our experimental design approach. 
    We are able to improve the error reduction in parameter estimation by $\sim 97\%$ compared to a uniform random sampling based method shown in Fig. (\ref{fig:block_param}), and generate more information-rich contacts as shown in Fig. (\ref{fig:block_fim}), with respect to the Fisher information parameters. 
    
    In general, we see better performance in our approach due to the higher excitations of tangent forces that produce gradients in our differentiable simulation that better identify parameters. We excite these gradients by generating sensor readings $\{ \bar{y}_i \}_{i=0}^N$ that maximizes terms in the score function $\xi_i \forall i \in [0,N]$ as described in (\ref{eq:FIM_general}), thus maximizing contact-aware Fisher information. 



\balance
\bibliographystyle{IEEEtran}
\bibliography{main}

\end{document}